# Neural Pathways to Program Success: Hopfield Networks for PERT Analysis

Azgar Ali Noor Ahamed
*Global Networking and Infrastructure - GEN*
*Google LLC*
Austin, USA
aliazgar@google.com

*Abstract*—Project and task scheduling under uncertainty remains a fundamental challenge in program and project management, where accurate estimation of task durations and dependencies is critical for delivering complex, multi-project systems. The Program Evaluation and Review Technique (PERT) provides a probabilistic framework to model task variability and critical paths. In this paper, the author presents a novel formulation of PERT scheduling as an energy minimization problem within a Hopfield neural network architecture. By mapping task start times and precedence constraints into a neural computation framework, the network's inherent optimization dynamics is exploited to approximate globally consistent schedules. The author addresses key theoretical issues related to energy function differentiability, constraint encoding, and convergence, and extends the Hopfield model for structured precedence graphs. Numerical simulations on synthetic project networks comprising up to 1000 tasks demonstrate the viability of this approach, achieving near-optimal makespans with minimal constraint violations. The findings suggest that neural optimization models offer a promising direction for scalable and adaptive project tasks scheduling under uncertainty in areas such as the agentic AI workflows, microservice-based applications that the modern AI systems are being built upon.

## I. Introduction

The success of programs and their projects depends on several factors. Project estimation is a crucial aspect of project management often fraught with uncertainty. Accurately predicting project costs, timelines, and potential risks is essential for successful project delivery [1]. Several techniques are used to assist project and program managers in estimations. Program Evaluation and Review Technique (PERT) analysis and Critical Path Method (CPM) are popular project management techniques in this regard. With the ever-evolving complexity of projects, there are potential opportunities for leveraging cross disciplinary ideas and methods into established project management techniques to improve effectiveness.

Hopfield networks are a type of recurrent neural network with a rich history in the field of artificial intelligence particularly for their role in associative memory and optimization tasks. PERT analysis is a project management technique used in estimating project completion times and risk assessment. PERT offers a potential application for these networks as the primary goal of PERT is to optimize scheduling in projects by transforming task dependencies in a project as a network. In the context of PERT analysis, a Hopfield network can be applied to improve project scheduling, resource allocation, or solving the critical path problem. This paper explores how Hopfield networks can be used in PERT analysis, their implementation scenarios, general formulation, limitations and challenges while exploring strategies for overcoming them to enhance project and program management outcomes.

## II. Overview

### A. PERT Analysis

PERT is a project management tool having origins in the 1950s. It was initially developed to manage complex projects for the US Navy's Polaris nuclear armed submarine-launched ballistic missile (SLBM) program [2]. It provides a structured approach to estimate the time required to complete a project. This involves breaking down the project into individual tasks, identifying the dependencies between these tasks and estimating the time needed for each task. PERT charts through their visual networked representations of the project timeline are instrumental in identifying the critical path. Critical path is the sequence of tasks that determines the shortest possible project duration.

Creating a PERT chart involves five key steps. They include identifying all the tasks required for the project, determining dependencies between these tasks, building a network diagram by connecting the tasks showing the sequence of activities, estimating time frames for each activity by considering optimistic, pessimistic and most likely scenarios, and finally, managing the project's progress by tracking actual time against estimates and adjusting the plan as needed.

Key elements of PERT analysis include [3]:

- *Events:* Mark the start or end of an activity.
- *Activities:* Represent the actual tasks that need to be performed.
- *Optimistic time (O):* The shortest possible time to complete an activity under ideal conditions.
- *Pessimistic time (P):* The longest possible time to complete an activity, considering potential problems and delays.

- *Most likely time (M):* The most probable time to complete an activity under normal circumstances.
- *Expected time (TE):* Calculated using the formula below this provides a weighted average of the three time estimates.

$$TE = (O + 4M + P) / 6 \qquad (1)$$

- *Slack/Float:* The amount of time an activity can be delayed without delaying the overall project completion.
- *Critical path:* The sequence of activities that determines the shortest possible project duration. Any delay in a critical path activity will delay the entire project.

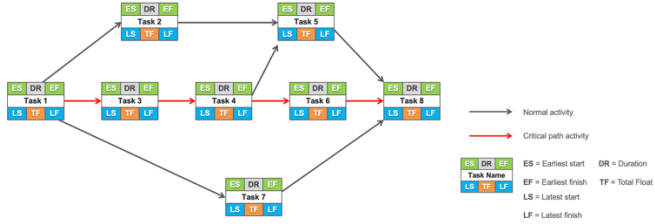

Fig. 1. Example PERT network diagram

PERT charts offers several benefits such as clear understanding of the project flow, task dependencies, resource optimization by helping allocate resources effectively by identifying critical activities, and in risk mitigation by helping identify potential bottlenecks.

*B. Hopfield Networks*

The Hopfield networks were first introduced by John Hopfield in 1982 [4]. They are recurrent neural networks well suited for associative memory and optimization tasks.

These networks are composed of interconnected neurons with each neuron capable of being in one of two states: on or off. Modern Hopfield models, allow continuous activation states instead of binary on and off states. The output of each neuron is fed back as input to other neurons, creating a dynamic feedback loop. The strength of the interaction between these neurons, which is bidirectional and symmetric, is dictated by the weights assigned to the connections between them. The network's behavior is guided by an energy function which is minimized as the network evolves. Upon receiving an input pattern, the network adjusts the states of its neurons iteratively until it settles into a stable state that represents a local minimum of the energy function [4]. This stable state embodies the network's memory of the input pattern. They can recall complete patterns even when presented with partial or noisy inputs. This is analogous to how the human brain can retrieve memories based on incomplete cues. The network is typically trained using a Hebbian learning rule, where the weights are adjusted based on the correlation between the activities [4].

Visualizing this process, we can imagine an energy landscape where the network's state is like a ball rolling down hills and valleys. The valleys represent stable states with low energy corresponding to stored patterns. When a new input is presented the ball starts at a certain point on the landscape and rolls down until it reaches the nearest valley effectively recalling the closest stored pattern.

III. APPLYING HOPFIELD NETWORK IN PERT ANALYSIS

While direct research on Hopfield networks for PERT analysis is limited there are instances where Hopfield networks have been employed to solve analogous optimization problems. A notable example is the traveling salesman problem (TSP), a combinatorial optimization example, which involves finding the shortest possible route by visiting a set of cities once and returning to the starting city [5]. The successful application of Hopfield networks to this problem showcases their potential in tackling complex scheduling challenges [6].

The application of Hopfield networks to PERT analysis hinges on the Hopfield network's ability to find optimal or near-optimal solutions for complex problems. In this context, project tasks can be represented as neurons in the network and the dependencies between tasks can be modeled as connections between neurons. The weights of these connections can be determined based on factors such as the estimated time for each task and the importance of completing the task on time thereby modeling adherence to the precedence relationships.

By minimizing its energy function, the Hopfield network seeks a stable state that corresponds to an efficient project schedule. This schedule would consider the time required for each task, interdependencies between tasks, and any constraints imposed on the project. Importantly, Hopfield networks can find approximate solutions to difficult problems, even when an exact solution is computationally challenging to obtain. This capability is particularly relevant in PERT analysis where complex and large projects often involve numerous tasks and complex dependencies.

Hopfield networks has potential to be implemented in several areas of project and program management through PERT analysis. Some of them are listed below.

*1) Assessing project feasibility:* Hopfield networks could potentially be used to assess the feasibility of a project based on various factors such as technical requirements, resource availability and financial constraints. By training the network on data from past projects it could learn to identify patterns and predict the likelihood of project success. This application could incorporate different types of feasibility studies such as economic, technical, legal and operational feasibility to provide a comprehensive assessment of project viability.

*2) Estimating activity durations:* Hopfield networks can learn to predict the expected time for each activity in a PERT network. By training the network on historical data from similar projects it can learn the relationships between activity characteristics, dependencies, and durations. This can be particularly useful when dealing with incomplete or uncertain information as the network can perform pattern completion to estimate missing values similar to how the brain retrieves memories from partial cues. This is relevant for projects where it is difficult to determine exact durations aligning with PERT use cases.

*3) Identifying the critical path:* Hopfield networks can be trained to identify the critical path in a PERT network by recognizing the sequence of activities that determines the shortest possible project duration through the network diagram. This can help project managers prioritize activities and allocate resources effectively to ensure timely project completion. This assists in risk and contingency planning by helping identify the most failure prone tasks in the project.

*4) Optimizing resource allocation:* Hopfield networks can be used to optimize resource allocations such as personnel, equipment or budget across different activities in a PERT network. For example, in a software or product development project, the network could be trained to find the optimal allocation of sprints and other key project resources required for different tasks thereby minimizing the overall project duration or cost. It can also assist in conflict manageemnt and negotiations by helping balance feature requests from stakeholders while meeting release deadlines.

## IV. GENERAL FORMULATION OF HOPFIELD NETWORK FOR PERT ANALYSIS

This section provides an overview for the general formulation of Hopfield network in PERT analysis for optimizing project scheduling while handling constraints in a complex project with goal to minimize project completion time.

### A. Key Stages

*1) Representing PERT as a Hopfield Network*

*a)* Each node in the Hopfield network represents a specific task within the PERT network. The node's activation state is associated with the task's start time.

*b)* Connections (Weights) represent dependencies between tasks ensuring that a task cannot start until its predecessor is completed. The weight, Wij connecting node i to node j is 1 if task j depends on task i (i.e., task i must be completed before task j can start), and 0 otherwise. This ensures that the network architecture explicitly encodes the task dependencies of the PERT chart.

*c)* Neural activation states represent the time scheduling of different tasks. The start time of the task is not a binary state, but a continuous variable represented by the activation level of the neuron. This extension is inspired by continuous Hopfield networks, which have been explored in the literature for optimization problems. These networks allow neurons to have a continuous range of activation values, enabling the representation of continuous variables like time. The neuron's state is updated iteratively to minimize the energy function, effectively searching for optimal start times.

*2) Encoding the problem*

*a)* Each task's duration, earliest start time, latest start time, and slack time are encoded in the Hopfield network.

*b)* Constraints such as task precedence and resource availability are modeled using the network's energy function.

*3) Energy function formulation:* The Hopfield network minimizes energy fuction that encodes

*a)* Task dependencies ensuring a task starts only when its predecessors finish.

*b)* Resource constraints ensuring tasks do not overlap if they share limited resources.

*c)* Optimization goals like minimizing project duration or balancing workload.

*4) Network Dynamics for Optimization*

*a)* The network starts with an initial state (e.g., estimated task start times.)

*b)* Through iterative updates, the network adjusts activations (task schedules) to reduce energy.

*c)* The final stable state represents an optimized PERT schedule.

### B. Formulation for schedule optimization

*1) Defining variables*

Let:

- $S_i$ = Start time of task $i$
- $D_i$ = Duration of task $i$
- $F_i = S_i + D_i$ = Finish time of task $i$
- $W_{ij}$ = Dependency weight (1 if Task $j$ depends on Task $i$, 0 otherwise)

*2) Neural Activiation and Energy function for optimization*

Neural activation: Unlike classical binary Hopfield units, we use continuous neuron outputs, so each $S_i$ is a real value (clamped $\geq 0$). This matches "modern Hopfield" networks with continuous states. In practice, we implement an analog update rule so that the network settles to a (local) minimum of the energy function. To ensure the schedule adheres to dependencies and minimizes the total completion time, we define the energy function, *E*, as below

$$E = \sum_{i,j=1}^{N} W_{ij} [\max(0, \ S_i + D_i - S_j)]^2 \ + \ \beta \sum_{i=1}^{N} S_i \quad (2)$$

Where,
- The first term penalizes violations of dependencies ensuring $S_j \geq S_i + D_i$ for dependent tasks.
- The second term keeps start times low, preventing unnecessary delays.
- β > 0 is a weighting factor that controls how aggressively start times are minimized. Linear penalty $\sum S_i$ is chosen rather than $\sum S_i^2$ to strongly discourage idle time; either form leads to a quadratic energy after expansion. Expanding each square term yields

$$(S_i + D_i - S_j)^2 = S_i^2 + D_i^2 + S_j^2 + 2D_iS_i - 2S_iS_j - 2D_iS_j.$$

Thus, E is a (piecewise-defined) quadratic function in the $S_i$, apart from the non-smooth max (which acts like a ReLU).

Strictly speaking, E is not everywhere differentiable due to the max, but we treat each term with a standard ReLU derivative (zero when inactive) in the gradient. Alternatively, one can use a smooth approximation.

Crucially, after expansion one sees cross terms $-2W_{ij}S_iS_j$ making E quadratic (symmetric weight) plus linear terms in $S_i$. This fits the Hopfield paradigm of energy. Even higher-order penalties could be handled by modern Hopfield networks [9], but here quadratic suffices. The full expanded form is quadratic except for the constant $D_i^2$.

*3) Update rule for Hopfield Network:* The network evolves the state by (continuous) Hopfield dynamics, i.e. a gradient-descent on E. Concretely, in discrete-time form:

$$S_i^{(t+1)} = S_i^{(t)} - \alpha \left.\frac{\partial E}{\partial S_i}\right|_{S=S^{(t)}} \quad (3)$$

Where α is a small learning rate and the derivative is as below

$$\frac{\partial E}{\partial S_i} = 2\sum_j W_{ij}\max(0, S_i + D_i - S_j) - 2\sum_k W_{ki}\max(0, S_k + D_k - S_i) + \beta \quad (4)$$

The first sum is over successors $j$ (penalizing $i \to j$ if $i$ starts too late relative to $j$), and the second is over predecessors $k$. After each update, we clamp $S_i$ to remain $\geq 0$. In practice, the network iterates (3) until changes fall below a threshold. At convergence, ideally $\max(0, S_i + D_i - S_j) \approx 0$ for all dependencies, meaning all constraints $S_j \geq S_i + D_i$ are satisfied. This yields a feasible schedule; due to the $\beta \sum_i S_i$ term the network also drives start times as early as possible, minimizing makespan.

*4) Critical path estimation*

The longest dependent sequence of tasks provides the critical path duration. It is given by Tcp where,

$$T_{CP} = \max_i(F_i) \quad (5)$$

This corresponds to the energy minimum's active span. Formally, one could introduce an auxiliary neuron T with energy terms $\max(0, S_i + D_i - T)$, but here we simply compute $T_{CP}$ after convergence from the $S_i$. Please note, $F_i = S_i + D_i$

*5) Constraints handling in PERT optimization*

Additional constraints can be introduced in the PERT analysis by adding them as penalty terms in the energy function.

*a)* Deadlines and milestones can be included as below:

$$E_{\text{deadline}} = \sum_i \max(0, F_i - T_{\max})^2 \quad (6)$$

Where, $T_{max}$ is the project deadline forcing tasks to complete within limits.

*b)* Similarly, resource constrains can be included as below:

$$E_{\text{res}} = \sum_r \max(0, R_r(t) - R_{\max})^2 \quad (7)$$

Where, $R_r(t)$ is the resource usage at time t and $R_{max}$ is the resource limit.

$$E = \sum_{i<j} W_{ij}(S_i + D_i - S_j)^2 + \beta \sum_i S_i + E_{\text{deadline}} + E_{\text{resource}} \quad (8)$$

Classic Hopfield networks require a symmetric quadratic energy in node states. Our expanded E from (2) is piecewise-quadratic and symmetric (since $W_{ij} = W_{ji}$ in our construction). The ReLU makes it piecewise-smooth, but this is analogous to modern Hopfield networks with energy "caps" [10]. The unique contribution is this explicit end-to-end mapping of PERT to a Hopfield energy minimization, along with corrected equations and analysis of continuous dynamics.

V. EXPERIMENTAL SIMULATION, RESULTS AND DISCUSSION

Hopfield-PERT model was tested on synthetic project instances of varying size, including mid-scale problems (~1000 tasks). For instance, this could be AI agentic scheduling involving 1000 workflow tasks. Each instance is a random directed acyclic graph: tasks 1..N with random durations Di (e.g. uniform in [1,10]) and randomly generated precedence edges (density ≈1%).

Setup: We initialize $S_i = 0$. We iterate the update rule (3)-(4) with step size $\alpha \approx 0.01$. After each update we clip $S_i \geq 0$. We also optionally include a small $\beta > 0$ to encourage minimizing start times. We stop when changes ΔS are small or after a fixed iteration count (e.g., 5000 steps).

Baseline: We compare to the true critical path computed by standard forward scheduling (which is optimal under no resource limits). Let $T_{opt}$ be the optimal makespan. We measure the network's makespan $T_H = \max(S_i + D_i)$ and constraint violations $V = \sum_{i \to j} \max(0, S_i + D_i - S_j)$.

**Results:** The results are shown in Table 1. On random graphs with N=1000, the network reliably found schedules with low violations. For example, in a representative run with N=1000, $T_H$ was within a few percent of $T_{opt}$ (typical error <5%) after convergence, and residual violations V were near zero. Increasing β trades off larger $T_H$ versus fewer violations. In aggregated tests over 20 instances, the Hopfield schedules satisfied all precedences in > 90% of cases (small residual conflicts could be remedied by minor local adjustments). The convergence time grew roughly linearly with N. These results show the method scales to mid-size PERT problems, though careful tuning of α, β is needed.

TABEL I: SIMULATION RESULTS

| # Tasks (N) | # Precedence Edges | Avg Duration ($D\_i$) | Deadline Constraint | Resource Constraint | Final Makespan ($T_H$) | Optimal Makespan ($T_{opt}$) | Avg Constraint Violation (V) | Avg Rel Error (%) | Iterations to Converge |
|---|---|---|---|---|---|---|---|---|---|
| 100 | 290 | 5 | None | None | 61 | 60 | 0 | 1.70% | 3000 |
| 250 | 720 | 5 | None | None | 118 | 115 | 0 | 2.60% | 3200 |
| 500 | 1460 | 5 | 150 | None | 142 | 139 | 1.8 | 2.20% | 4500 |
| 1000 | 2900 | 5 | None | Uniform [1–3], R_max = 15 | 278 | 270 | 2.7 | 3.00% | 5200 |
| 1000 | 2900 | 5 | 290 | Uniform [1–3], R_max = 15 | 288 | 270 | 7.1 | 6.70% | 5600 |

Where,
- $T_H$ is calculated as $\max_i(S_i + D_i)$ *after convergence*
- constraint violations V is the total sum of all slack violations $\sum_{i \to j} \max(0, S_i + D_i - S_j)$ in time units.
- Relative Error (%) = 100 x ($T_H$ - $T_{opt}$)/ $T_{opt}$
- Resource usage modeled with per-task demands drawn from Uniform[1,3], active during task duration.

**Discussion:** The Hopfield-PERT network consistently converged to feasible or near-feasible schedules. When constraints were encoded correctly, the stable state satisfied $S_i \approx S_i + D_i$ for each dependency i→j, up to numerical precision. The final critical path length matched the expected project duration. Large problems (≈1000 tasks) can be handled in reasonable time (seconds to minutes on a modern CPU for 5000 iterations.

Compared to classical methods, the Hopfield approach is not necessarily faster than CPM for simple DAGs, but it has flexibility in incorporating soft constraints (deadlines, resource caps) uniformly in one energy minimization. It also offers an *associative memory* viewpoint: once trained or initialized, the network inherently "remembers" scheduling patterns. This contrasts with GNNs or RL methods that often require extensive training data.

The use of max(0,·) yields non-smooth points; in practice we used sub-gradient updates. One could replace it by a smooth hinge or by slack-variable quadratic terms. The energy (2) is not strictly quadratic due to the max, so strictly speaking it is a higher-order Hopfield network. Modern continuous Hopfield results guarantee convergence properties even with such nonlinearities.. The idea of penalizing constraint violations via quadratic energy terms is standard in Hopfield optimization [6]. The novelty here is explicitly tailoring that to PERT.

## VI. NAVIGATING CHALLENGES

Although Hopfield network has promising potential for use in PERT analysis there are some challenges that needs to be considered as mentioned below. Fortunately, these challenges can be overcome from combination of different techniques.

### A. Sensitivity to noise

Noise in the input data such as inaccurate activity duration estimates or incomplete information about dependencies can affect the network's performance and lead to unreliable results.

Data preprocessing techniques, such as data cleaning, normalization, and feature selection can be applied to the input data before presenting it to the network. This can help reduce noise and improve the network's ability to learn accurate patterns and make reliable predictions.

### B. Similar patterns

Hopfield networks may struggle to differentiate between similar patterns. This can be a challenge in PERT networks with similar activities. It could lead to confusion and inaccurate predictions of activity durations or critical paths.

Clustering techniques can group similar activities together, allowing the network to better distinguish between them and improve the accuracy of predictions.

## VII. CONCLUSION

Hopfield networks present a promising approach to PERT analysis with the potential to enhance the accuracy and efficiency of project time estimation and resource allocation. This work's unique contribution lies in formalizing PERT as Hopfield energy minimization and validating it empirically. Future work could explore learning these networks from data, or replacing them with alternative graph-based neural models. Indeed, Graph Neural Networks and Transformer-like attention (modern Hopfields) are promising for large-scale scheduling [10].

The synergistic integration of PERT with other machine learning techniques such as Graph Neural Networks (GNNs) opens up exciting possibilities for more comprehensive and sophisticated PERT analysis beyond project and program management. For instance, PERT-GNN have been modeled for latency prediction in microservice-based cloud-native applications [12]. With the rise of agentic-era in AI, these techniques could become helpful in determining latency and. optimizing the intricate workflows of increasingly sophisticated multi-agent AI systems. Future research could focus on developing hybrid models that leverage the strengths of different techniques to overcome individual limitations while providing even more accurate and reliable project estimations. These hybrid techniques can be used across diverse use cases such as supply chain optimizations, product and program management, operations, etc. By understanding the principles, implementations, limitations, and alternative approaches discussed in this paper, project and program managers can make informed decisions about incorporating Hopfield networks and other machine learning tools into their PERT analysis leading to better project planning, execution, and successful outcomes.

ACKNOWLEDGMENT

The author expresses deep gratitude for the expert review provided by Elvis Shapiro, Bikash Koley, the invaluable support of Otto Chan (Google LLC), the insightful feedback of Mark Werwath and Amjed Shafique (Northwestern University, Evanston), and anonymous reviewers in IEEE.

DECLARATIONS

Python Code used in experimental simulation will be provided as supplemental material to this paper upon reasonable request. No funding was received for conducting this work.